\documentclass[runningheads]{llncs}
\usepackage[T1]{fontenc}
\usepackage{graphicx}
\usepackage{subfig}
\usepackage{hyperref}
\usepackage{booktabs}
\usepackage{subcaption} 

\usepackage{marvosym}
\usepackage{multirow}
\usepackage{makecell}

\usepackage{bbding}

\begin{document}
\title{Patch Progression Masked Autoencoder with Fusion CNN Network for Classifying Evolution Between Two Pairs of 2D OCT Slices}
\titlerunning{AMD Progression Classification in MICCAI 2024}%
%
%
\author{
Philippe~Zhang\inst{1,2,3} \and
Weili~Jiang\inst{4}\and
Yihao~Li\inst{5}\and 
Jing~Zhang\inst{1,2}\and
Sarah~Matta\inst{1,2}
Yubo~Tan\inst{6}\and
Hui~Lin\inst{7}\and
Haoshen~Wang\inst{8}\and
Jiangtian~Pan\inst{9}\and
Hui~Xu\inst{10}\and
Laurent Borderie\inst{3}\and
Alexandre Le Guilcher\inst{3}\and
Béatrice Cochener\inst{1} \and
Chubin~Ou\inst{11}\and
Gwenolé Quellec\inst{1}\and 
Mathieu Lamard\inst{1,2}
}
\authorrunning{P. Zhang et al.}
%
\institute{LaTIM UMR 1101, Inserm, Brest, France  \and
University of Western Brittany, Brest, France \and
Evolucare Technologies, Villers-Bretonneux, France \and
College of Computer Science, Sichuan University, Chengdu, China \and
United Imaging Healthcare, Shanghai, China\and
University of Electronic Science and Technology of China, Chengdu, China \and
Northwestern University, Evanston, IL, USA \and
ShanghaiTech University, Shanghai, China \and
Wuhan University, Wuhan, China\and
MD Anderson Cancer Center, Houston, USA\and
Guangdong General Hospital,Guangzhou, Guangdong, China
}

\maketitle              
\begin{abstract}
Age-related Macular Degeneration (AMD) is a prevalent eye condition affecting visual acuity. Anti-vascular endothelial growth factor (anti-VEGF) treatments have been effective in slowing the progression of neovascular AMD, with better outcomes achieved through timely diagnosis and consistent monitoring. Tracking the progression of neovascular activity in OCT scans of patients with exudative AMD allows for the development of more personalized and effective treatment plans. This was the focus of the Monitoring Age-related Macular Degeneration Progression In Optical Coherence Tomography (MARIO), in which we participated. In Task 1, which involved classifying the evolution between two pairs of 2D slices from consecutive OCT acquisitions, we employed a fusion CNN network with model ensembling to further enhance the model’s performance. For Task 2, which focused on predicting progression over the next three months based on current exam data, we proposed the Patch Progression Masked Autoencoder that generates an OCT for the next exam and then classifies the evolution between the current OCT and the one generated using our solution from Task 1. The results we achieved allowed us to place in the \textbf{Top 10} for both tasks. Some team members are part of the same organization as the challenge organizers; therefore, we are not eligible to compete for the prize. The code is available at 
\url{https://github.com/pzhangwj/mario_challenge_code}.

\end{abstract}

\section{Introduction}
Age-related Macular Degeneration (AMD) is a common condition affecting vision in individuals over 65 \cite{jonas2017updates}. Since 2007, anti-VEGF treatments \cite{rosenfeld2006ranibizumab} have been vital for managing neovascular AMD, relying on early diagnosis and consistent monitoring \cite{rasmussen2014long,freund2015treat}. Developing effective computer-assisted diagnostic tools for AMD remains a key research focus \cite{li2020treatment}. Optical coherence tomography (OCT) \cite{lains2021retinaloct} is crucial for diagnosing and monitoring AMD, guiding treatment decisions.

Deep learning (DL) has transformed medical imaging, significantly advancing the detection of eye conditions like AMD \cite{bhuiyan2020artificial} and Diabetic Retinopathy (DR) \cite{dai2021deepdiabetic}. Notable DL models for time-series prediction include Long Short-Term Memory (LSTM) \cite{hochreiter1997lstm} and Vision Transformer (ViT) \cite{dosovitskiy2020vit}, which effectively capture temporal and spatial patterns, respectively, enhancing disease progression prediction \cite{ashir2021diabeticlstm,holste2024harnessingvit}.

The MARIO Challenge\footnote{\url{https://youvenz.github.io/MARIO_challenge.github.io/}} aims to evaluate algorithms for detecting changes in neovascular activity from OCT scans in AMD patients. The challenge includes two tasks: analyzing consecutive OCT B-scan images to classify changes and predicting disease progression from a single scan. Our team achieved top-10 results by employing the Fusion CNN Network and a Vision Transformer-based method for these tasks.

\section{Materials and methods}

\subsection{Dataset Description}

The MARIO dataset was collected to monitor the progression of AMD in 136 patients from Europe. For disease monitoring, each patient undergoes an examination using the Spectralis OCT device (Heidelberg Engineering) with follow-up option on. Each examination includes a set of successive 2-D OCT slices (B-scans) that form a 3-D OCT volume, a 2-D infrared image (localizer) corresponding to each 3-D volume, and associated data such as age, sex, visit number, and eye laterality. The dataset is divided as follows for both tasks: 68 patients for training, 34 for validation, and 34 for testing. To maintain the challenge’s integrity and fairness, the validation and test datasets are processed by the organizers to ensure an unbiased evaluation. The test dataset used for final ranking is securely held and not released to participants.\\

The first Task (Classify evolution between two pairs of 2-D slices from two consecutive 2-D OCT acquisitions) is focused on a four-category image classification. The categories are as follows:
\begin{itemize}
    \item Reduced (0): The condition has been classified as reduced, indicating that it has either been eliminated or is persistently reduced.
    \item Stable (1): The condition is considered stable, meaning it is either inactive or persistently stable.
    \item Worsened (2): The condition has worsened, signifying that it is either persistently worsened or has newly appeared.
    \item Other (3): The condition falls into the "other" category, which includes cases that are uninterpretable or have both appeared and then been eliminated.
\end{itemize}

The second task (Prediction of AMD progression within 3 months using OCT 2D slices) focuses on classifying the progression into the same categories as in Task 1, excluding the "Other" category (3).\\

The data annotation was meticulously performed by ophthalmologists specializing in retina care, each with at least two years of experience in monitoring vascular AMD patients. Initially, seven labels were used, but this was simplified to four categories for Task 1 and three for Task 2, as described above. For the training and validation cases, a single ophthalmologist handled the annotation. In contrast, two ophthalmologists independently annotated the test cases, with the agreement between their annotations serving as a baseline for evaluating algorithm performance.

\subsection{Data processing}
\begin{figure}
    \centering
    \includegraphics[width=0.8\linewidth]{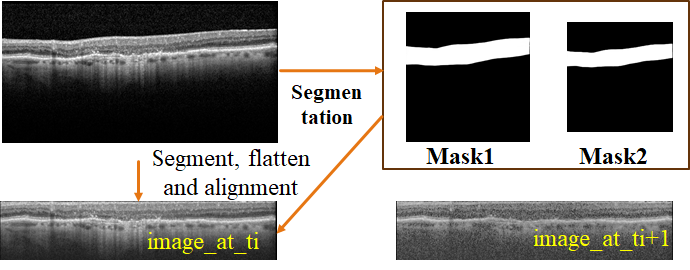}
    \caption{Optical Coherence Tomography Image Preprocessing}
    \label{fig:preprocessing}
\end{figure}

OCT slices often contain significant noise and irrelevant information. While the Follow-Up option allows for precise horizontal realignment of image pairs, vertical (depth) alignment remains inconsistent. To address this and improve comparability between image pairs, we applied Optical Coherence Tomography Image Preprocessing(OCTIP\footnote{\url{https://github.com/leto-atreides-2/octip}})\cite{Thomas}, developed by LaTIM Labs (\url{https://latim.univ-brest.fr/}). Several Feature Pyramid Network (FPN) architectures from the EfficientNet family were trained for the segmentation task, and among them, FPN-EfficientNet-B6 and FPN-EfficientNet-B7 achieved the best performance. We used these two top-performing models to generate segmentation masks and applied the median of their outputs to ensure a robust prediction, minimizing the impact of potential classification errors or over-segmentation from a single model. Once the segmented regions are extracted, we align the upper boundary of the retina (inner limiting membrane) with the top of the image, effectively "flattening" the retina and realigning the images along the depth axis. Additionally, by eliminating the vitreous, it enables us to zoom in on the retina, providing more detailed information to the neural network, ensuring more accurate and consistent analysis. The processing flow is illustrated in Figure \ref{fig:preprocessing}.

\subsection{Task 1}
\begin{figure}
    \centering
    \includegraphics[width=0.9\linewidth]{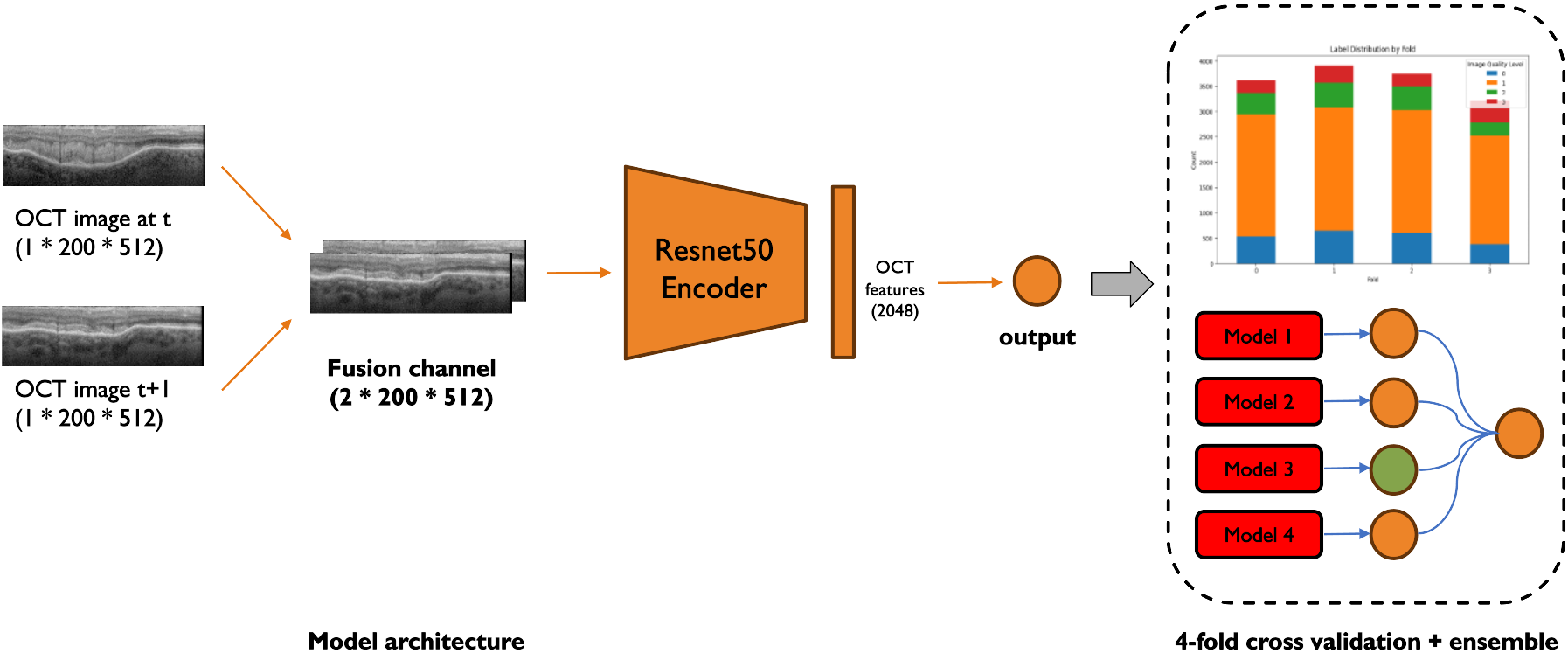}
    \caption{Early Fusion CNN Network }
    \label{fig:2}
\end{figure}

In Task 1, we explored two fusion methods \cite{li2024review}: Early Fusion and Late Fusion networks. During the experimentation phase, we tested various backbones, and ResNet50 showed the best performance.

\subsubsection{Early Fusion Network}

Illustrated in Figure \ref{fig:2}, this architecture concatenates OCT images along the channel dimension. 
Each OCT image, with a size of \(1 \times 200 \times 512\), is combined to form a fused input of size \(2 \times 200 \times 512\). This fused input is passed through a ResNet50 encoder pre-trained on ImageNet, which extracts feature maps of size 2048. The extracted features are then processed through fully connected layers to perform the final classification task.

\subsubsection{Late Fusion Network}

\begin{figure}
    \centering
    \includegraphics[width=0.9\linewidth]{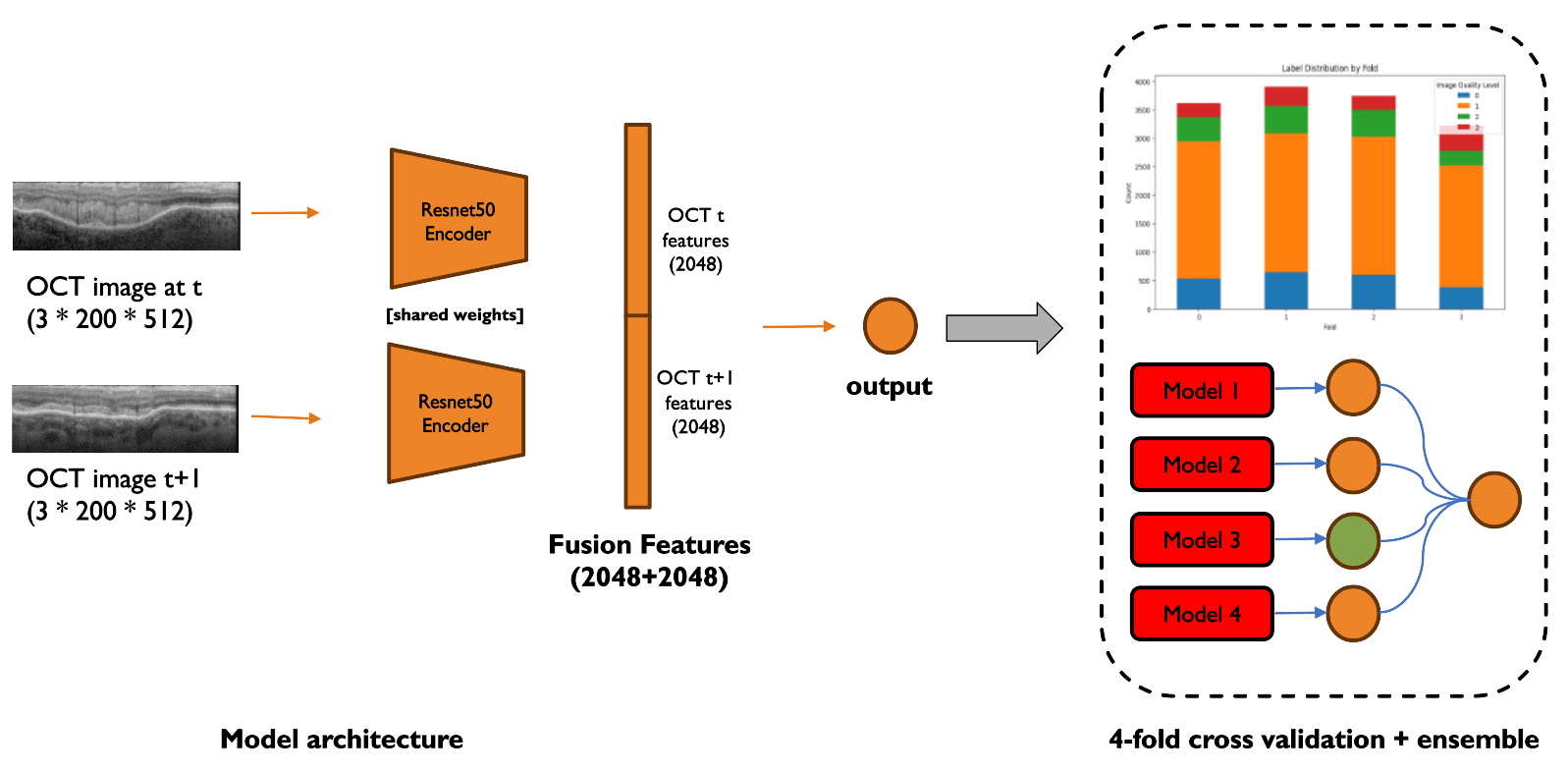}
    \caption{Late Fusion CNN Network }
    \label{fig:1}
\end{figure}

As shown in Figure \ref{fig:1}, this architecture fuses image features from two different time points. The input consists of OCT images at time \(t\) and \(t+1\) (with dimensions of 3\(\times\)200\(\times\)512). Since OCT images are originally grayscale, they are repeated three times to simulate RGB format input. The OCT images from two different time points are fed into a ResNet50 encoder to extract 2\(\times\)2048-dimensional features. The features extracted from the two time points are then concatenated to form a 4096-dimensional fusion feature vector (2048+2048), which is then passed through a fully connected layer to produce the classification output.

Additionally, we adopted a 4-fold cross-validation strategy, training four models on different folds (Model 1 to Model 4) and combining their results through ensembling to further improve the model's robustness and generalization\cite{zhang2024detection}.

\subsection{Task 2}

\begin{figure}
    \centering
    \includegraphics[width=0.95\linewidth]{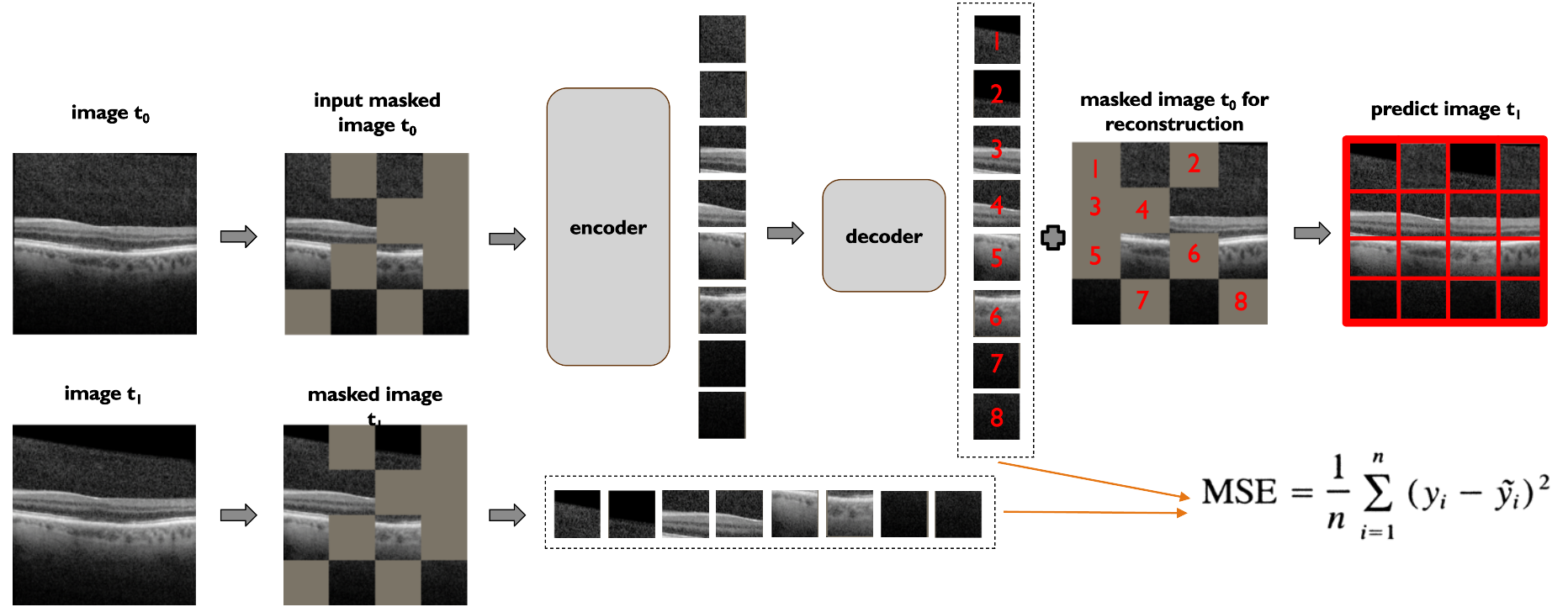}
    \caption{Vision Transformer Patch Progression Masked AutoEncoder }
    \label{fig:3}
\end{figure}

Task 2 focuses on classifying disease progression based on a single examination. To address this, we propose a novel method called Patch Progression Masked Autoencoder (PPMAE), inspired by the Masked Autoencoder (MAE) \cite{he2022masked}. While Task 1 provides two scans to classify disease progression, Task 2 only supplies one scan. Therefore, we train PPMAE on the Task 1 dataset to predict a future OCT image based on the current scan. By generating this predicted image, we can apply the classification model from Task 1 to assess disease progression, effectively using both the current and predicted future scans to determine progression based on two scans.

\subsubsection{Patch Progression Masked Autoencoder}

Illustrated in Figure \ref{fig:3}, we mask 75\% of the baseline OCT image at time \(t_0\), similar to the traditional MAE method. However, instead of reconstructing the same image from the unmasked part, PPMAE predicts the patches of the follow-up image at time \(t_1\). The model learns to use the masked patches of the \(t_0\) image to predict the corresponding progression patches from the \(t_1\) image. This strategy allows the model to capture temporal changes between the patches, learning the disease progression over time.

The predicted patches are then aligned with the unmasked regions of the image \(t_0\), resulting in a reconstruction of the complete future OCT image that corresponds to the image several months later. The reconstruction error is calculated using the Mean Squared Error (MSE) between the predicted patches and the true patches from the follow-up image. This approach not only enables the model to predict the future state of the disease but also helps in improving classification performance for progressive cases by leveraging temporal information.

Finally, we leverage our classification model from Task 1 to perform the disease progression classification, using the reconstructed \(t_1\)
  image along with \(t_0\) to complete the task.

\subsection{Implementation~Details}
The implementation details for our experiments are summarized in Table~\ref{tab:4}. All our models were trained on a NVIDIA A6000 with 48GB of memory.

\begin{table}[htb]
\centering
\caption{Implementation~details used in experiments.}
\label{tab:4}
\scalebox{1.0}{
\resizebox{\linewidth}{!}{
\begin{tabular}{c|c|c|c}
\hline  
\textbf{Implementation} & \textbf{Task1 } & \textbf{Task2 Finetuning} & \textbf{PPMAE}   \\
\hline
\textbf{Input size} & \multicolumn{2}{c|}{512 $\times$ 200 pixels}  &  224 $\times$ 224 pixels  \\
\hline
\textbf{Backbone} &\multicolumn{2}{c|}{ResNet50}  & MAE ViT Large \\
\hline
\textbf{Library} & \multicolumn{3}{c}{timm} \\
\hline
\textbf{Pretrained weights} &\multicolumn{2}{c|}{Imagenet}& Imagenet MAE\\
\hline
\textbf{Loss} &\multicolumn{2}{c|}{CrossEntropyLoss} &  MSELoss \\
\hline
\textbf{Optimizer} &  \multicolumn{3}{c}{AdamW} \\
\hline
\textbf{Learning rate} & 1e-4 (w/o scheduler) & 1e-5 (w/o scheduler) & 1e-5 (w/o scheduler) \\
\hline
\textbf{Augmentation} & \multicolumn{2}{c|}{\parbox{9cm}{RandomHorizontalFlip, RandomVerticalFlip,  RandomRotation,\\ ColorJitter, RandomPerspective, GaussianBlur}} & \parbox{3.5cm}{RandomResizedCrop, \\ RandomHorizontalFlip}\\
\hline
\textbf{Batch size} &\multicolumn{3}{c}{128} \\
\hline
\textbf{Epochs} & 150 & 1 & 100 \\
\hline
\textbf{Train/Val split} & 0.75:0.25 & 1:0 & 0.75:0.25\\
\hline
\textbf{Metric} & Accuracy, F1, Specificity, RkC &Accuracy, F1, RkC, Specificity, QWK &  MSE \\
\hline
\end{tabular}}}
\end{table}

\section{Results}

Table \ref{tab:1} shows OCTIP improves performance for both Early Fusion and Late Fusion methods in Task 1 of the MARIO Challenge. OCTIP enhanced F1 scores and rank correlation, particularly benefiting both Late Fusion method.

\begin{table}[ht]
\centering
\caption{Preliminary results on MARIO Challenge - Task1}\label{tab:1}
\scalebox{1}{
\begin{tabular}{c c c c c}
\toprule  
Method & OCTIP & F1 score & Rk correlation & Specificity \\
\midrule 

Early Fusion & \tiny{\XSolid} & 0.8037 & 0.5790 & 0.8857\\
Early Fusion & \tiny{\Checkmark} & 0.8181 & 0.6094 & 0.8894\\
Late Fusion & \tiny{\XSolid} & 0.8130 &0.6070 & 0.8967 \\
Late Fusion & \tiny{\Checkmark} & 0.8223 &0.6270 & 0.9013 \\
\bottomrule 
\end{tabular}}
\end{table}

\begin{table}[h]
\centering
\caption{Reconstruction Accuracy (MSE) on Task 1 Validation Set}\label{tab:2}
\scalebox{1}{
\begin{tabular}{c c c c }
\toprule  
Method & Preprocess (OCTIP) & MSE \\
\midrule MAE & \tiny{\XSolid} & 0.5572\\
PPMAE & \tiny{\XSolid} & 0.3269 \\
PPMAE & \tiny{\Checkmark} &0.1724\\
\bottomrule 
\end{tabular}}
\\(In MAE, we follow the traditional training approach by masking parts of the image and attempting to reconstruct the masked patches. The goal remains to reconstruct the image at time \(t_1\).)\\
\end{table}

Table \ref{tab:2} shows the image \(t_1\) reconstruction performance, the PPMAE model achieved a MSE of 0.3269 without preprocessing and 0.1724 when preprocessing was applied. In comparison, the traditional MAE resulted in a higher MSE of 0.5572, indicating that the PPMAE model significantly improves reconstruction accuracy, especially when preprocessing is applied.  

\begin{figure}
    \centering
    \includegraphics[width=1\linewidth]{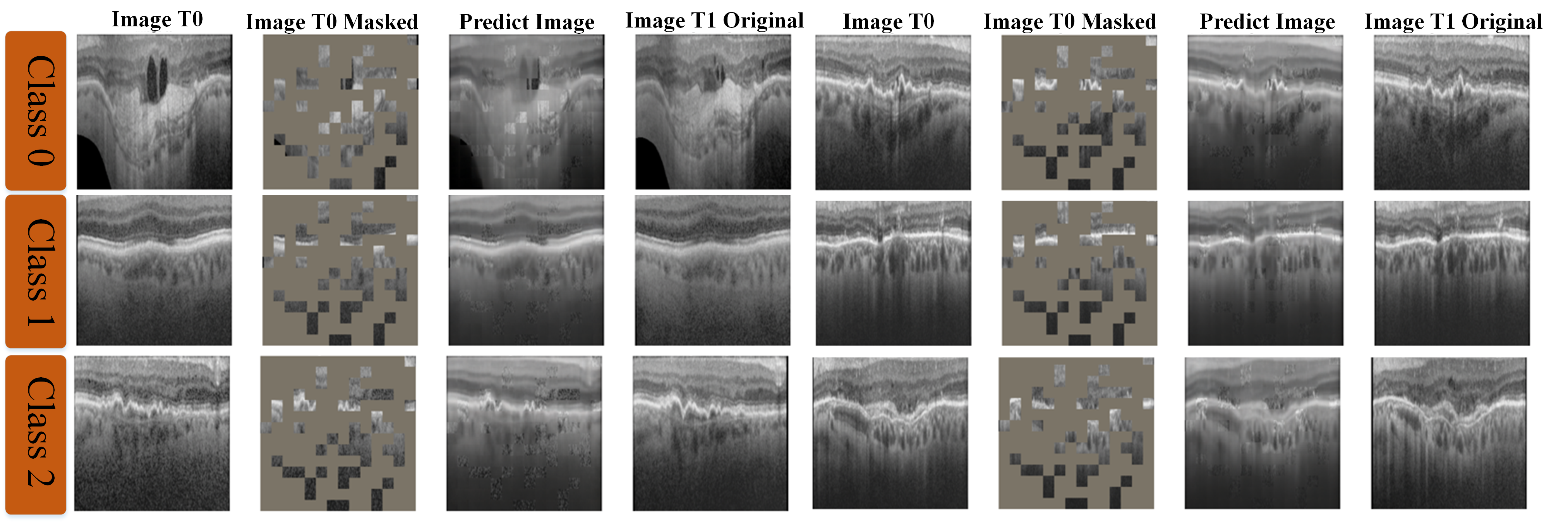}
    \caption{Visual Comparison of OCT Reconstruction Results by PPMAE}
    \label{fig:ppmae_result}
\end{figure}

The reconstruction results of the three categories are shown in Fig. \ref{fig:ppmae_result}. As shown in the figure, the model is still able to produce high-quality predictions and maintain good structural similarity with the original images.

\begin{table}[h]
\centering
\caption{Preliminary results on MARIO Challenge - Task2}\label{tab:3}
\scalebox{0.8}{
\begin{tabular}{c c c c c c} 
\toprule  
Method & F1 score & Rk correlation & Specificity &  WQK & Mean metrics \\
\midrule 
MAE + Task 1 models & 0.6465 & 0.0790 & 0.6936 & 0.0060 & 0.3563 \\ 
PPMAE + Task 1 models & 0.6491 & 0.0938 & 0.6963 & 0.0410 & 0.3701 \\ 
PPMAE + Task 1 models finetuned on data Task 2 & 0.6933 & 0.0923 & 0.6910 & 0.0227 & 0.3748 \\ 
\bottomrule 
\end{tabular}}
\\(In MAE, we follow the traditional training approach by masking parts of the image and attempting to reconstruct the masked patches. The goal remains to reconstruct the image at time \(t_1\).)\\
\end{table}

Table \ref{tab:3} shows that the PPMAE-based method performs better across multiple metrics. This indicates that a better reconstruction significantly enhances performance. The approach using PPMAE with finetuning achieves the best overall performance, with a F1 score of 0.6933 and a mean metric value of 0.3748. Meanwhile, the approach using PPMAE without finetuning outperforms the other methods in terms of Rk correlation and weighted quadratic kappa. Additionally, in MAE, the reconstruction of \(t_1\) often fails, with the reconstructed image resembling \(t_0\) more than \(t_1\) , which results in most predicted classes being stable.

\section{Discussion and Conclusions}

In this work, we presented our solution for the two tasks in the MARIO Challenge. For Task 1, we utilized Late Fusion CNN Network with ResNet50 as the encoder, combined with OCTIP, to achieve the best results in feature extraction and classification accuracy. Leveraging model ensembling further enhanced our performance metrics.  For Task 2, we introduced the PPMAE, which predicts future OCT images based on a single exam, demonstrating strong reconstruction and classification performance. Notably, OCTIP played a key role in both tasks by improving image quality, enhancing classification accuracy, and enabling more precise and efficient feature extraction, which also contributed to better image reconstruction. Our methods yielded excellent results, placing us in the Top 10 for both tasks. These outcomes highlight the potential of advanced neural network architectures in developing more personalized and effective treatment strategies for AMD.

However, predicting disease progression from a single exam remains challenging. Adding patient data, such as age, sex, visit number, and eye laterality, did not improve performance. This aligns with our observation that these factors are poorly correlated with disease progression labels. Further research into image reconstruction methods, particularly for generating better future images, is essential to continue improving performance in this field.

\begin{credits}
\subsubsection{\discintname}The authors have no competing interests.
\end{credits}

\bibliographystyle{splncs04}
\newpage 

\bibliography{ref}

\end{document}